# Iterative Deep Convolutional Encoder-Decoder Network for Medical Image Segmentation

Jung Uk Kim, Hak Gu Kim, and Yong Man Ro*, *Senior Member, IEEE*

*Abstract—* In this paper, we propose a novel medical image segmentation using iterative deep learning framework. We have combined an iterative learning approach and an encoder-decoder network to improve segmentation results, which enables to precisely localize the regions of interest (ROIs) including complex shapes or detailed textures of medical images in an iterative manner. The proposed iterative deep convolutional encoder-decoder network consists of two main paths: convolutional encoder path and convolutional decoder path with iterative learning. Experimental results show that the proposed iterative deep learning framework is able to yield excellent medical image segmentation performances for various medical images. The effectiveness of the proposed method has been proved by comparing with other state-of-the-art medical image segmentation methods.

## I. INTRODUCTION

Medical image segmentation is a necessary task of identifying and localizing malignancy or particular anatomic structures in medical images for diagnosis, planning, treatment, etc. For the segmentation, there are manual approach, semi-automatic (interactive) approach, and fully automatic approach. The manual segmentation approach requires time-consuming and labor-intensive process. The semi-automatic approach requires an expert experience and a prior knowledge so that it could limit practical applications. As such, it is needed to develop a fully automatic medical image segmentation that automatically extracts regions of interest (ROIs) in medical images.

A variety of automatic medical image segmentation methods have been proposed for a body part or lesion region such as vessel and skin cancer. For robust segmentation in dental X-ray images, hand-craft feature-based automatic segmentation method was proposed. Pedro et al. proposed a segmentation method using shape model generation [1]. In [1], the optimized shape model was used to extract features and boundary information for distinguishing tooth region from background regions. In [2], infinite perimeter active contour model was used to perform automated retinal vessel segmentation. Authors of [3] proposed multi-scale classification-based lesion segmentation method for dermoscopic images. In [3], they incorporated collection of multi-scale classifiers to distinguish between lesion and surrounding skin parts.

As seen in the previous works, it is difficult to design generic hand-craft features for various segmentation cases. In particular, hand-craft features or statistic model designed for a body part or a lesion could not be directly applied to other medical image segmentation.

Recently, deep learning has attracted significant attention because it shows high performance due to deep latent features of raw input data. Lower layers containing low level features feed into higher layers. Higher layers capture global information by using filters which contains broader receptive fields. By combining these hierarchical features, deep learning framework could represent robust feature for medical segmentation such as lung tissue, brain tumor segmentation and so on [14]. In [4], convolutional neural network (CNN)-based biomedical image segmentation was proposed in a fully automated fashion. Authors of [4] proposed the U-shaped network architecture for a precise localization. However, the loss of resolution (details of segmentation) in the convolution process could happened and it could not be compensated enough in the up-sampling layers. In addition, one end-to-end learning and resultant segmentation could not correctly segment complex textural regions seen in typical medical images.

In this paper, we propose a new medical image segmentation method using deep learning which takes into account details of segmentation. The main contributions of our paper are twofold:

1) This paper proposes a novel convolutional encoder -decoder network for an effective medical segmentation. In our network, we combine the low-level features of a medical image learned at the encoder part and segmented image at the decoder part by a copy connection. So the deconvolution layers followed by the encoder part with the copy connection could lead to accurate and detail segmentation results.

2) Iterative learning in the proposed deep network is devised for detail segmentation of medical image. Medical images are likely to include complex boundary and low contrast region so that medical image segmentation performance could be degraded. To improve medical image segmentation performance, in this paper, interim segmentation result is iteratively fed to the input with a given medical image. By iteratively minimizing errors between the interim segmentation results and ground-truth, the proposed network can achieve high segmentation performance.

Furthermore, to avoid over-fitting by limited database in medical domain, we employ data augmentation and transfer learning. Experimental results show that the proposed method achieves better performances than existing methods for various medical image segmentation tasks including lesion segmentation of skin cancer and retinal vessel segmentation.

The rest of the paper is organized as follows. In the section

J. U. Kim, H. G. Kim, and Y. M. Ro are with the Image and Video Systems Lab., School of Electrical Engineering, Korea Advanced Institute of Science and Technology, Republic of Korea (e-mail: jukim0701@kaist.ac.kr, hgkim0331@kaist.ac.kr, ymro@ee.kaist.ac.kr) *:corresponding author 2, we describe the proposed medical image segmentation method. In the section 3, experiment results are presented to verify the performance of the proposed method. Finally, conclusions are drawn in the section 4.

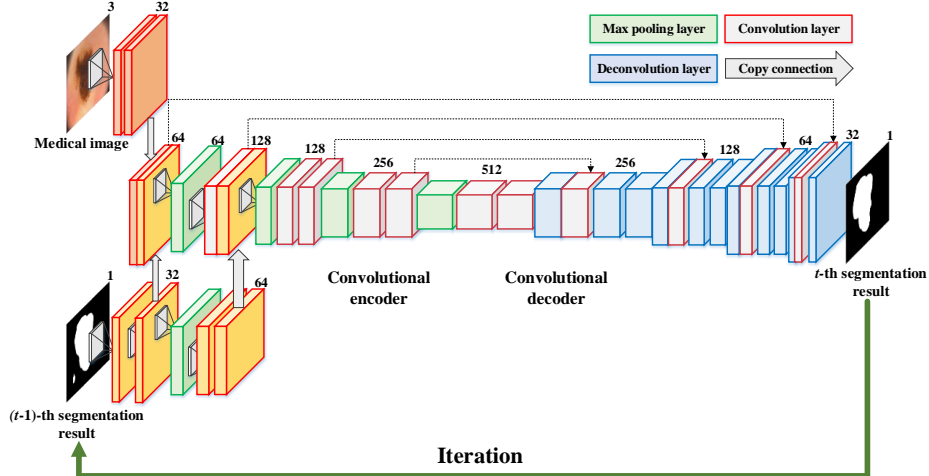

Fig. 1. Overall procedure of the proposed iterative convolutional encoder-decoder network for medical image segmentation at the *t*-th iteration step. Note that the input of the network is a medical image and the (*t*-1)-th iteration segmentation image. The output is the segmentation result.

## II. METHOD

Fig. 1 shows the overall procedure of the proposed iterative convolutional encoder-decoder network (at the *t*-th iteration step) for the medical image segmentation. As shown in the figure, the proposed network consists of convolutional layers for encoder and deconvolution layers for decoder. Furthermore, we devise an iterative encoder-decoder network to iteratively improve segmentation result. To that end, both the medical image and interim segmentation result are iteratively fed to the proposed network, as shown in the Fig. 1. Detailed descriptions of each part are given in the following sections.

### A. Deep convolutional encoder-decoder network

As shown in Fig. 1, the proposed network consists of two main parts: convolutional encoder part and decoder part. The convolutional encoder part consists of convolution layers and max-pooling layers. In our network, the encoder part requires two inputs. One is the original medical image and the other is (*t*-1)-th interim segmentation result. The input resolutions are $256 \times 320$ in our experiment. In this paper, to encode reliable features related to the ground truth of segmentation, we combine feature maps from the medical image and those from the interim segmentation, as shown in Fig. 1. For this, the feature maps of the second convolution layer are made by concatenating copies from the two inputs (called as copy connection. Please see the first layer of the encoder part in Fig. 1). In addition, the feature map of the fourth layer with the (*t*-1)-th interim segmentation result is concatenated with the feature map of corresponding layer in the encoder network, which helps to precisely localize detail region to be segmented. By combining the feature maps of both the medical image and interim segmentation result, segmentation result is refined. In the encoder network, filters with size of $3 \times 3$ are learned in convolution layer. The maximum pool layer of the $2 \times 2$ window reduces the resolution of the feature map by half. As the feature map resolution goes half, the channel number of the feature map is doubled. By convolution and pooling operations, encoder features are learned with the two input images (original medical image and interim segmentation result).

The decoder part consists of deconvolution layers. The encoded features of the convolutional encoder part are fed to the convolutional decoder part to predict target regions be segmented. In the decoder network, filters with size of $3 \times 3$ deconvolution operations are learned in every deconvolution layer. Due to the max pooling process, detail information in the original image could be lost. To mitigate the loss of detail information, skip connection method is adopted. As shown in Fig. 1 (dotted arrow lines), convolution layers and deconvolution layers are linked by concatenating feature maps of the encoder and corresponding feature maps of the decoder. In contrast to the convolutional encoder part, the feature map resolution of the decoder part goes double as the channel number of feature map goes half. At the final layer, $1 \times 1$ convolution with sigmoid activation is used. Consequently, we obtain a binary segmentation result with size of $256 \times 320$.

### B. Iterative deep learning network for refinement

The medical image segmentation is challenging due to noise, complex texture, and unclear boundaries [5]. The convolutional encoder-decoder network could predict the target region to be segmented. However, complex textures or detailed structures of the target region in medical images could not be predicted due to the down-sampling layers in the convolutional encoder parts. In this paper, we devise the iterative approach with the deep convolutional encoder-decoder network for the segmentation in the medical images. To that end, a given medical image and the output of the previous iteration (i.e., interim binary segmentation result) are fed to the inputs of the proposed network for the next iteration, as shown in Fig. 1. After that, the segmentation result is iteratively updated by minimizing errors between the predicted result and ground-truth.

At the first iteration, the initial segmentation map is fed to the network. The initial value for the initial segmentation map is set to 0.5, which is between zero (background region) and one (segmentation region). In the following step, the previous segmentation result is fed to the input of the network. Therefore, in the proposed iterative convolutional encoder-decoder network, the segmentation result can be improved iteratively.

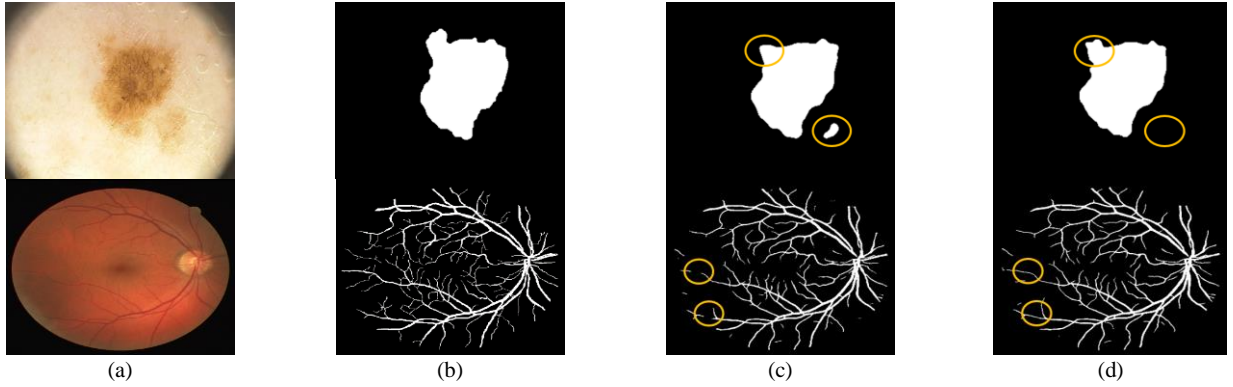

Fig. 2. Examples of the segmentation result for PH2 and DRIVE. The first row and second row are examples for PH2 and DRIVE datasets, respectively. (a) Original image, (b) Ground-truth of the segmentation, (c) Result of the first iteration, (d) Result of the final iteration.

In training stage, segmentation result and the proposed network parameter are iteratively updated. The iteration is terminated if the difference between the outputs of the previous and current iteration is small enough, which can be written as

$$\sum_{i=1}^{P}\left|S_i^t - S_i^{t-1}\right| < Th, \quad (1)$$

where $S_i^t$ and $S_i^{t-1}$ denote the *i*-th pixel values of segmentation result at the *t*-th and (*t*-1)-th iteration, respectively. *Th* is the threshold. *P* is the total number of pixels of the segmentation map.

### C. Objective function for iterative convolutional encode-decode network

In proposed iterative approach, we devise a new objective function based on dice coefficient. Dice coefficient is a metric for measuring the similarity of two samples. With this metric, the overlap ratio between ground-truth and predicted output can be measured. At the *t*-th iteration step, dice coefficient between the ground-truth and segmentation result is written as

$$D_t = \frac{2\sum_{i=1}^{P} y_i \hat{y}_i^t}{\sum_{i=1}^{P}(y_i)^2 + \sum_{i=1}^{P}(\hat{y}_i^t)^2}, \quad (2)$$

where $y_i$ and $\hat{y}_i^t$ denote the *i*-th pixel value of the ground-truth and predicted segmentation image at the *t*-th iteration step, respectively. *P* is the total number of pixels.

The proposed objective function at *t*-th iteration, $L_t$, can be written as

$$L_t(\theta_t) = -\frac{D_t + \varepsilon}{D_{t-1} + \varepsilon}, \quad (3)$$

where $\varepsilon$ is a constant value to avoid instability. $\theta_t$ is the network parameters of the *t*-th iteration step. $D_{t-1}$ and $D_t$ is the dice coefficient values at the (t-1)-th and t-th iteration step, respectively.

By using (3), the proposed network is regularized to draw the network parameters of *t*-th iteration, $\theta_t$, which increases the dice coefficient between the ground-truth and segmentation result at the *t*-th iteration. Namely, if the dice coefficient of the (*t*-1)-th iteration is low, the gradient of the objective function becomes large to increase $D_t$. On the other hand, if the dice coefficient of the (*t*-1)-th iteration is high, the gradient of the objective function becomes small to converge $D_t$.

## III. EXPERIMENTS AND RESULTS

### A. Datasets

To evaluate the performance of the proposed method, we performed experiments with two public medical datasets: PH2 [6] and DRIVE [7]. The PH2 dataset was used for melanocytic lesion segmentation. It contains 200 dermoscopic images of melanocytic lesions. Due to the limited number of samples, the data augmentation was conducted to avoid over-fitting. In our experiment, each image was horizontally and vertically flipped, rotated in 9 types of degrees ([-16°, 16°] with an increment of 4°) and translated in 25 types of ranges (±40, ±20 and 0 in x-y axis). As a result, 900 augmented images from each image were generated. In our experiment, 36,000 images were used for training. 160 dermoscopic images were used for test. The DRIVE dataset with 40 retinal images was used for retinal vessel segmentation. We also performed the data augmentation. Each image was flipped horizontally and vertically, rotated in 13 types of degrees ([-24°, 24°] with an increment of 4°) and translated in 25 types of ranged (±20, ±10 and 0 in x-y axis). 1,300 images were generated from each image. In our experiment, 26,000 images were used for training. 20 retina images were used for test. In addition, we pre-trained our network using transmission electron microscopic (TEM) dataset [13] which contained 1,294 cell images. By augmentation, we generated 46,584 images which were used for pre-training our network. After the pre-training, we performed fine-tuning of our network with PH2 and DRIVE.

### B. Performance evaluation of medical image segmentation

To show the effectiveness of the proposed iterative network, segmentation was performed with different iteration steps. Fig. 2 shows the segmentation results of the proposed network at the first and last iteration. In Fig. 2, the first and second rows were the results of lesion segmentation for PH2 and those of retinal vessel segmentation for DRIVE, respectively. Fig. 2 (a) and (b) show the original and ground-truth images, respectively. Fig. 2 (c) and (d) show the results of the first and the final iteration, respectively. In Fig. 2 (b)-(d), white area and black area indicate segmented region (value is 1) and other regions (value is 0), respectively. As shown in Fig. 2 (c), wrong parts were detected at the first iteration (shown by yellow circles). On the other hand, as shown in the Fig. 2 (d), the wrong detected part was removed and detailed structures were segmented at the final iteration (shown by yellow circles).

TABLE I. PERFORMANCE COMPARISON FOR SEGMENTATION.

| Methods | PH2 DC | PH2 JC | Methods | DRIVE DC | DRIVE JC |
|---|---|---|---|---|---|
| AT [8] | 0.805 | 0.720 | AT [8] | 0.574 | 0.403 |
| Active Contours [9] | 0.853 | 0.760 | Multi-scale top-hat [11] | 0.730 | 0.574 |
| Abedini [3] | 0.910 | 0.840 | Liu [12] | 0.769 | 0.625 |
| SSLS [10] | 0.913 | 0.847 | IPACHI [2] | 0.782 | - |
| U-net [4] | 0.924 | 0.854 | U-net [4] | 0.795 | 0.672 |
| **Proposed Method** | **0.940** | **0.882** | **Proposed Method** | **0.826** | **0.708** |

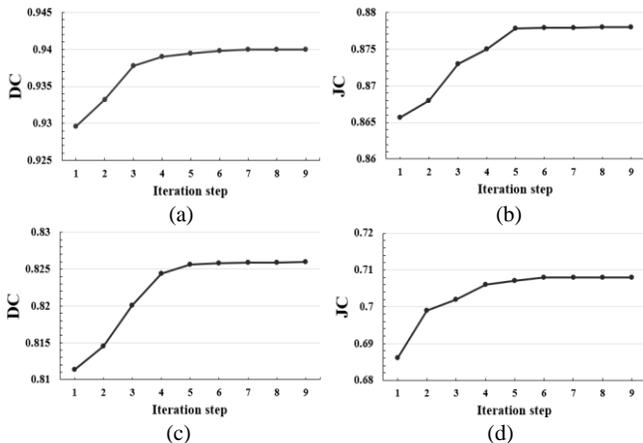

Fig. 3. DC and JC according to iteration steps for PH2 and DRIVE. (a) DC result for PH2, (b) JC result for PH2 dataset. (c) DC result for DRIVE, (d) JC result for DRIVE.

For the assessment of segmentation performance, we measured two metrics [3]: dice coefficient (DC) and Jaccard coefficient (JC). Note that DC and JC measure the overlap ratio of the ground-truth and predicted segmentation result. Table I shows the segmentation performance measured by DC and JC for PH2 and DRIVE. In PH2 dataset, for performance evaluation, we compared with five state-of-the-art methods: four methods were hand-craft based models; one was a deep learning based model. In hand-craft based models, AT [8], Level Set Active Contours [9], Abedini's Method [3] and SSLS [10] were used. In a deep learning based model, U-net [4] was used. As shown in Table I, the results of the proposed method were 0.940 of DC and 0.882 of JC for lesion segmentation.

In the DRIVE dataset, we compared with five state-of-the-art methods: four methods were hand-craft based models; one method was a deep learning based model. In hand-craft models, AT [8], Multi-scale top-hat [11], Liu's method [12], and IPACHI method [2] were used. In a deep learning based model, U-net [4] was also used. As shown in Table I, the proposed method achieved 0.826 of DC and 0.708 of JC for vessel segmentation. In both datasets, the proposed methods outperformed the five previous methods.

Furthermore, we verified the effectiveness of the iterative approach of the proposed network. We measured DC and JC between the ground-truth and iterative segmentation result according to the iteration. Fig. 3 shows DC and JC in iteration steps for PH2 and DRIVE datasets. Fig. 3 (a) and (b) show DC and JC for PH2 dataset, respectively. Fig. 3 (c) and (d) show DC and JC for DRIVE dataset, respectively. As shown in Fig. 3, DC and JC increase as iteration increase until converged. Consequently, experimental results demonstrated that the proposed iterative deep learning-based segmentation improved the performance of medical image segmentation.

## IV. CONCLUSIONS

In this paper, we proposed a new medical image segmentation using iterative deep encoder-decoder network. In the encoder part, hierarchical features were encoded using convolution and max-pooling layer of both original medical image and interim segmentation result. In the decoder part, we adopt deconvolution layer and skip connection to predict the target region to be segmented. We proposed iterative approach and new objective function to improve detail segmentation performance of the medical image. The experimental results showed that the proposed method outperformed existing medical image segmentation method.

## V. REFERENCES

[1] P. H. M. Lira, G. A. Giraldi, and L. A. P. Neves, "Panoramic dental x-ray image segmentation and feature extraction," in *Proc. V Workshop of Computing Vision*, 2009.
[2] Y. Zhao, L. Rada, K. Chen, S. P. Harding, and Y. Zheng, "Automated vessel segmentation using infinite perimeter active contour model with hybrid region information with application to retinal images," *IEEE Trans. Med. Imag.*, vol. 32, no. 9, pp. 1797-1807, 2015.
[3] M. Abedini, et al. "Multi-scale classification based lesion segmentation for dermoscopic images," in *Proc. IEEE Int. Conf. Engineering in Medicine and Biology Society*, 2016.
[4] O. Ronneberger, P. Fischer, and T. Brox, "U-net: convolutional networks for biometrical image segmentation," in *Proc. Int. Conf. Medical Image Computing and Computer-Assisted Intervention*, 2015.
[5] B. N. Li, C. K. Chui, S. Chang, and S. H. Ong, "Integrating spatial fuzzy clustering with level set methods for automated medical image segmentation," *Computers in biology and medicine*, vol. 41, no. 1, pp. 1-10, 2011.
[6] T. Mendonca, P. M. Ferreira, J. S. Marques, A. R. S. Marcal, and J. Rozeira, "Ph2-a demoscopic image database for research and benchmarking," in *Proc. IEEE Int. Conf. Engineering in Medicine and Biology Society*, 2013.
[7] J. J. Staal, M. D. Abramoff, M. Niemeijer, M. A. Viergever, and B. V. Ginneken, "Ridge based vessel segmentation in color images of the retina," *IEEE Trans. Med. Imag.*, vol. 23, no. 4, pp. 501-509, 2004.
[8] M. Silveira, J. C. Nascimento, J. S. Marques, A. R. S. Marcal, R. Mendonca, S. Yamauchi, J. Maeda, and J. Rozeira, "Comparison of segmentation methods for melanoma diagnosis in dermoscopy images," *IEEE J. Sel. Topics Signal Process*, vol. 3, no. 1, pp. 35-45, 2009.
[9] T. F. Chan, B. Y. Sandberg, and L. A. Vese, "Active contours without edges for vector-valued images," *Journal of Visual Communication and Image Representation*, vol. 11, no. 2, pp. 130-141, 2000.
[10] E. Ahn, L. Bi, Y. H. Jung, J. Kim, C. Li, M. Fulham, and D. D. Feng, "Automated saliency-based lesion segmentation in dermoscopic images," in *Proc. IEEE Int. Conf. Engineering in Medicine and Biology Society*, 2015.
[11] M. Liao, S. W. Zheng, and Y. Q. Zhao, "A novel method for retinal vascular image enhancement," *Journal of Optoelectronics Laser*, vol. 23, no. 11, pp. 2237-2242, 2012.
[12] L. Liu, T. Yang, D. Fu, and M. Li, "Retinal vessel extraction and diameter calculation based on tensor analysis," *in Society of Instrument and Control Engineers of Japan*, 2016.
[13] V. Morath, et al. "Semi-automatic procedure for the determination of the cell surface area used in systems immunology," *Frontiers in Bioscience, Elite*, vol. 5, pp. 533-545, 2013.
[14] G. Hayit, B. V. Ginneken, and R. M. Summers, "Guest editorial deep learning in medical imaging: overview and future promise of an exciting new technique," *IEEE. Trans. Med. Imag.*, vol. 35, no. 5, pp. 1153-1159, 2016.